\newif\ifcameraready
\camerareadyfalse
\newif\ifarxiv
\arxivtrue
\ifcameraready\ifarxiv
  \errmessage{Set only one of arxiv and cameraready to true}
\fi\fi

\newif\ifieeeconfmissing
\ifcameraready
  \newcommand{\loadpaperclass}{%
    \documentclass[letterpaper, 10pt, journal, twoside]{IEEEtran}}
\else
  \IfFileExists{ieeeconf.cls}{%
    \newcommand{\loadpaperclass}{%
      \documentclass[letterpaper, 10pt, conference]{ieeeconf}%
      \IEEEoverridecommandlockouts  
      \overrideIEEEmargins}         
  }{%
    \ieeeconfmissingtrue
    \newcommand{\loadpaperclass}{%
      \documentclass[letterpaper, 10pt, conference]{IEEEtran}%
      \IEEEoverridecommandlockouts}
  }
\fi
\loadpaperclass
\ifieeeconfmissing
  \newenvironment{keywords}{\begin{IEEEkeywords}}{\end{IEEEkeywords}}
\fi

\usepackage{cite}
\usepackage{amsmath,amssymb,amsfonts}
\usepackage{graphicx}
\usepackage{booktabs}
\usepackage{multirow}
\usepackage{makecell}      
\usepackage{xcolor}
\usepackage{url}
\usepackage{balance}       
\usepackage[hidelinks]{hyperref}
\ifcameraready\else\ifarxiv\else\hypersetup{pdfauthor={}, pdfcreator={}}\fi\fi

\graphicspath{{figures/}}

\newcommand{\todo}[1]{\textcolor{red}{[TODO: #1]}}

\newcommand{\sysname}{Curie's Jarvis}   

\newcommand{\paperkeywords}{AI-Enabled Robotics, Task Planning, Perception
for Grasping and Manipulation, Robotics and Automation in Life Sciences}

\ifcameraready
\fi

\begin{document}
\bstctlcite{IEEEexample:BSTcontrol}

\ifcameraready
\else
\title{\LARGE \bf \sysname: Autonomous Robotic Chemistry Experiments from
Printed Protocols}
\fi

\ifcameraready
\else\ifarxiv
  \author{Rumi Loghmani$^{*}$, Bozhang Song$^{*}$, Victor Alvarado,
  Aliyah Redding, and Amir Barati Farimani%
  \thanks{$^{*}$These authors contributed equally.}%
  \thanks{The authors are with the Department of Mechanical Engineering,
  Carnegie Mellon University, Pittsburgh, PA 15213, USA.}}
\else
  \author{Anonymous Author(s)}
\fi\fi

\maketitle
\ifcameraready\else
  \thispagestyle{empty}
  \pagestyle{empty}
\fi
\ifieeeconfmissing
  \noindent\todo{ieeeconf.cls not found, so this is IEEEtran conference
  format, not the RA-L submission format. Put ieeeconf.cls next to
  main.tex (on Overleaf: upload it to the project).}
\fi

\begin{abstract}
A person running a bench experiment works from a printed instruction
sheet, tells identical containers apart by the labels written on them,
and looks at the bench again before each step. Lab automation usually
works the other way around: the procedure is written as a script, and
containers stand at fixed positions or carry fiducial markers. We present
\sysname{}, a system that lets a robot arm carry out a chemistry
experiment on a bench set up for people. It reads the steps from a photo
of the printed sheet, finds containers on an ordinary bench with no fixed
positions or markers, tells identical-looking containers apart by reading
their labels in four camera views, locates its target again before every
step, and sets how each step is carried out, including its speed, from
what is being handled. On a Franka Panda arm, \sysname{} completed 5 of 5
runs of one protocol from a children's chemistry kit and 4 of 5 runs of
another, with no human intervention, and in 160 locates on printed and
handwritten labels it identified every container correctly. This brings
robots a step closer to doing bench chemistry the way people do.
\end{abstract}

\ifcameraready
\else
\fi

\section{Introduction}
\label{sec:intro}
A student running a bench protocol works from a printed instruction sheet.
The reagents sit in containers that have been labelled by hand and set down
wherever there was room, and identical cups are told apart only by their
labels. The bench is shared, so a cup may not be where it was a minute ago;
the student looks again before each step and finds it. The student also
adjusts each step to what is being handled, carrying an empty cup quickly and
a full one slowly. None of this is arranged for a robot: the instructions,
the labels and the layout are all meant for people. 

Existing robots for automated experiments mostly adapt the experiment's
procedure, objects and layout to the system. The instructions have to be
written as a script or in a chemical programming
language~\cite{steiner2019organic,mehr2020universal,fakhruldeen2022archemist};
every container and instrument needs a fixed or known
position~\cite{burger2020mobile,dai2024autonomous}; and some systems
identify containers by fiducial markers such as AprilTags rather than by
the labels placed with them~\cite{yoshikawa2022chemistry}. This makes
automated experiments fast and accurate, but at the cost of flexibility: a
new experiment requires substantial manual reprogramming instead of simply
handing the system a printed instruction sheet, and if an object is moved
during the experiment, the run may not be able to continue.

This work aims to close this gap by letting a robotic system adapt to a lab
arranged for people. We make this goal concrete as five properties. Our system,   \sysname{}, (P1)~starts from the printed instruction sheet, with no manual
rewriting into a script; (P2)~works on an ordinary bench, where containers
need no fixed or preset positions and no visual markers; (P3)~tells
identical-looking containers apart by their hand-written or printed labels;
(P4)~looks at the bench again before every step, so it finds objects that
have been moved; and (P5)~adapts how each step is executed (how fast to move, how far to tip, how deep to dig) to what is handled and what the step asks for. Section~\ref{sec:related}
compares related work along these properties (Table~\ref{tab:related}), and
Section~\ref{sec:experiments} evaluates each of them.

\begin{figure}[!t] \centering \includegraphics[width=.96\linewidth]{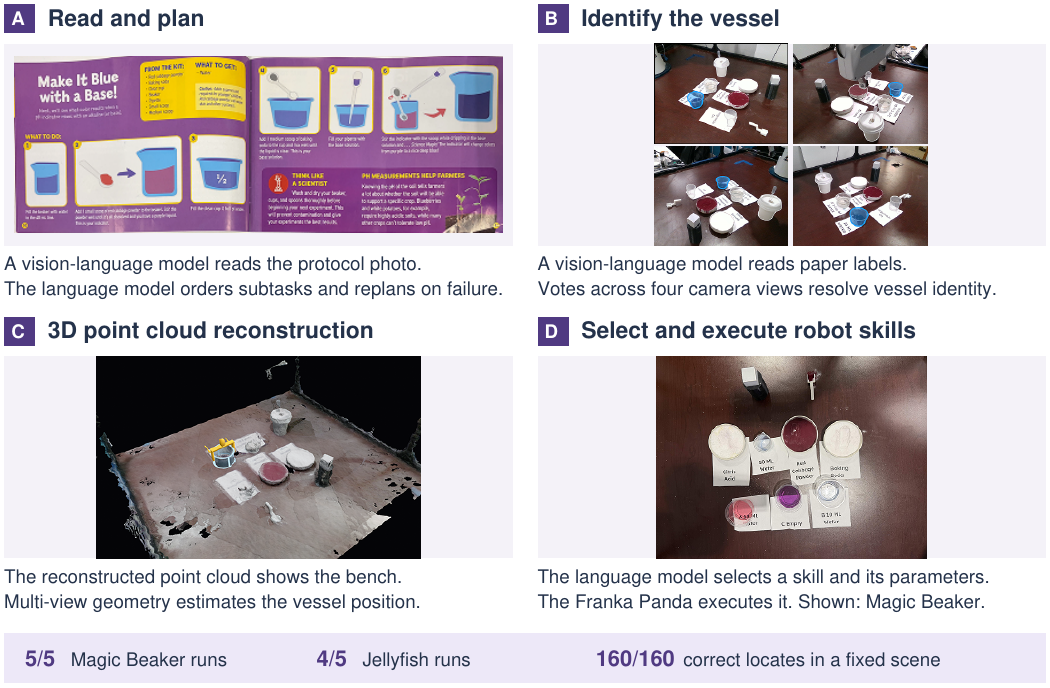} \caption{\sysname{} workflow. (A) A vision-language model reads the protocol photo. The LLM orders subtasks and replans after failures. (B) Votes across four camera views identify vessels from paper labels. (C) A 3D point cloud reconstructs the bench; multi-view geometry locates the target vessel. (D) The LLM selects predefined skills and their parameters for the Franka Panda. Shown: Magic Beaker.} \label{fig:teaser} \end{figure}

\begin{figure*}[t]
  \centering
  \includegraphics[width=\textwidth]{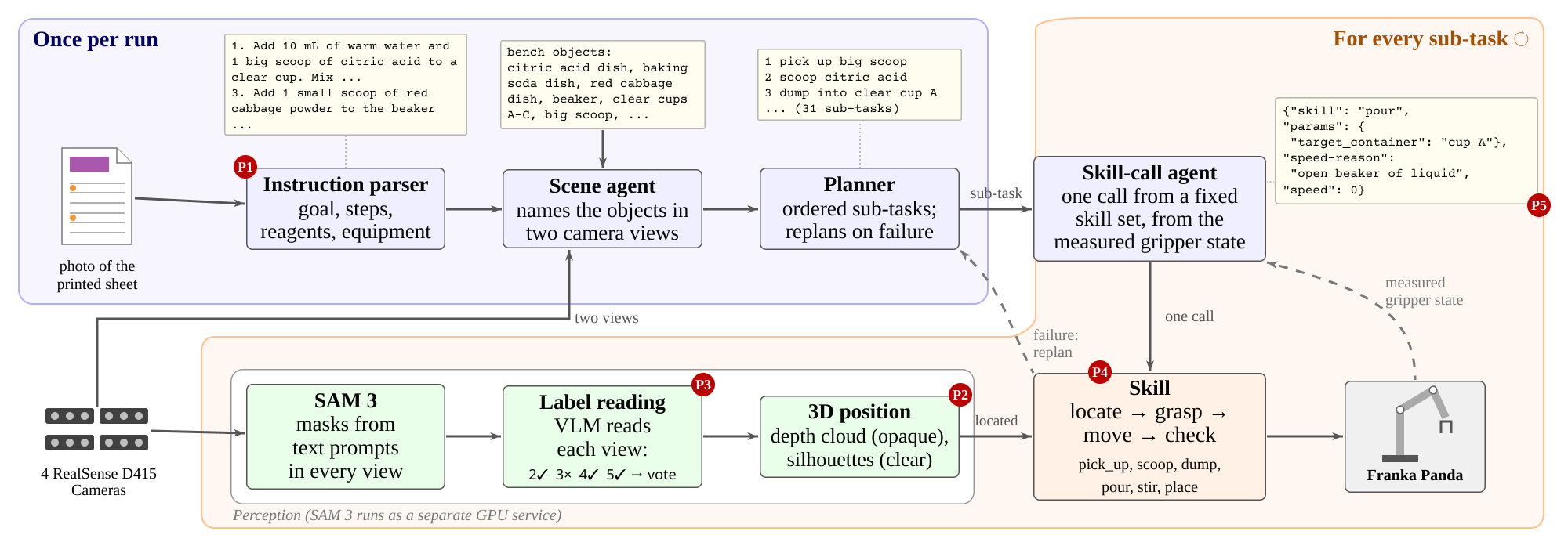}
  \caption{System architecture of \sysname{}. Once per run, the instruction
parser reads the steps from a photo of the sheet, the scene agent names the
objects on the bench, and the planner orders the sub-tasks. For each
sub-task, the skill-call agent produces one skill call, and the skill locates
its target again before it acts. Red markers show where P1--P5 are
implemented.}
  \label{fig:architecture}
\end{figure*}

We implement \sysname{} with all five properties on a Franka Panda arm
(Fig.~\ref{fig:teaser}). (P1)~A vision-language model (VLM) extracts the
steps from a photo of the sheet, and a planner turns them into parameterized
calls to a fixed set of skills. (P2)~SAM~3~\cite{carion2026sam} segments
objects in four camera views, and positions come from multi-view geometry:
the depth point cloud for opaque objects and silhouettes for transparent
ones. The system only needs a list of the objects on the bench (with the
size of each transparent container), not their positions. (P3)~The VLM
reads each container's label in every view, and the cameras vote.
(P4)~Before each skill call, the target is located again if it has been
handled or its image has changed. (P5)~The VLM chooses each call's pour
angle from the wording of the step and its speed from what is held; a fixed
rule sets the angles when the same container is poured from, or into, more
than once, and the scoop depth comes from the measured powder surface.

The contributions of \sysname{} are as follows:
\begin{itemize}
  \item A system that carries out a chemistry experiment on a Franka Panda
  arm from a photo of the printed instruction sheet, on an ordinary bench
  with no fixed positions or markers for the containers. It completed 5 of
  5 Magic Beaker runs and 4 of 5 jellyfish runs with no human
  intervention.
  \item A way to tell identical-looking containers apart by their
  hand-written or printed labels, which they stand on, read in each of four
  camera views and combined by a cross-camera vote. Because the position of
  each container comes from multi-view geometry, the label only needs to say
  what the container is and can be placed loosely. In 160 locates on
  printed labels and the handwriting of three people, every container was
  identified correctly.
  \item Re-locating and re-identifying the containers before every step,
  so the robot can find a container a person has moved between steps,
  even among identical-looking ones.
\end{itemize}

\section{Related Work}
\label{sec:related}

\textbf{Robotic chemists and lab automation.}
Much of the reliability of automatic laboratories~\cite{tom2024self}
comes from adapting the experiment to the robot system. Procedures are written as scripts or in a chemical programming language~\cite{steiner2019organic,mehr2020universal,fakhruldeen2022archemist};
robots work between fixed stations, with fixed sample racks and taught
paths~\cite{dai2024autonomous,lunt2024modular}, and one localizes at each
station by touching a calibration cube to within
$\pm$0.12\,mm~\cite{burger2020mobile}; and containers or devices are
identified or localized with fiducial markers such as AprilTags or QR
codes~\cite{yoshikawa2022chemistry,angers2025roboculture,vescovi2023towards}.
More recently, large language models generate procedures from natural
language and pass them to liquid handlers or cloud
platforms~\cite{boiko2023autonomous,m2024augmenting} or to robot arms at
the bench~\cite{yoshikawa2023large,darvish2025organa,wei2026agentchemist,zhang2025robochemist}. The closest to our system is
ORGANA~\cite{darvish2025organa}: it also uses a Franka arm and perceives
transparent glassware without markers, but the user assigns the role of each
identical-looking vessel at start-up, and it does not replan online. As
Table~\ref{tab:related} shows, none of these systems starts from a printed
instruction sheet, tells containers apart by reading their labels, or is
evaluated with a person moving objects mid-run.

\textbf{Language models for robot planning.}
The role of language models in robotics has grown from turning
instructions into skill sequences, programs or
trajectories~\cite{ahn2022can,singh2022progprompt,liang2023code,huang2023voxposer}
to setting execution parameters and closing the loop. For execution
parameters, Code as Policies~\cite{liang2023code} and
IROSA~\cite{knauer2026irosa} adjust speed in response to a user's
natural-language corrections, and ELLMER~\cite{mon2025embodied} selects contact
forces from object properties. For closed-loop operation, Inner
Monologue~\cite{huang2022inner} re-detects the scene after every step, and
BioProVLA-Agent~\cite{du2026bioprovla} uses a VLM to check conditions before
and after each sub-task. Our system does not leave execution entirely to the
model: the model chooses the pour angle from the wording of the step and the
speed from what is held, a fixed rule assigns the angles when the same
container is poured from, or into, more than once, and the scoop depth comes
from the measured powder surface.

\textbf{Identifying objects in the lab.}
Fiducial markers such as AprilTags~\cite{olson2011apriltag,wang2016apriltag}
give both an exact identity and a pose, but every object needs its own tag,
and the offset from each tag to its object must be calibrated, which is
fragile and time-consuming to maintain~\cite{angers2025roboculture}.
Open-vocabulary models~\cite{liu2024grounding,carion2026sam} need no
markers, but they find objects by category and cannot tell identical-looking
containers apart. Robots have identified objects by reading the text printed
on them with OCR~\cite{sharma2023semantic,liu2021robot,khan2025shake}. We
read the hand-written or printed label that each container stands on and
combine the reads across four views; the label only says what the container
is, while its position comes from multi-view geometry, so the same label can
be read by both people and the robot.

 \begin{table}[!ht]
  \caption{Comparison along the properties of Sec.~\ref{sec:intro}:
  P1 instruction input, P2 no fixed positions or markers, P3 container
  identification, P4 re-perception before each step, P5 adapted execution.
  \checkmark: yes; $\circ$: partial; --: no or not reported.}
  \label{tab:related}
  \centering
  \footnotesize
  \begin{tabular}{@{}lcccccc@{}}
    \toprule
    System & P1 & P2 & P3 & P4 & P5 & Platform \\
    \midrule
    Chemputer~\cite{steiner2019organic,mehr2020universal} & script & -- & -- & -- & -- & Fluidic \\
    Burger et al.~\cite{burger2020mobile}       & script & -- & --      & --      & --      & Mobile arm \\
    Coscientist~\cite{boiko2023autonomous}      & text   & -- & --      & --      & --      & Liquid handler \\
    CLAIRify~\cite{yoshikawa2023large}          & text   & $\circ$ & markers & $\circ$ & --      & Arm \\
    ORGANA~\cite{darvish2025organa}             & text   & $\circ$ & user    & --      & --      & Arm \\
    AgentChemist~\cite{wei2026agentchemist}     & text   & $\circ$ & --      & $\circ$ & $\circ$ & Dual-arm \\
    RoboChemist~\cite{zhang2025robochemist}     & text   & $\circ$ & --      & $\circ$ & $\circ$ & Dual-arm \\
    \midrule
    \sysname{} (ours)                           & photo  & \checkmark & labels & \checkmark & \checkmark & Arm \\
    \bottomrule
  \end{tabular}
\end{table}

\section{System Overview}
\label{sec:overview}

\textbf{Hardware.} 

The system runs on a Franka Emika Panda arm with a Franka Hand
parallel-jaw gripper. Four Intel RealSense D415 RGB-D cameras are mounted
on a cage around the bench and view it from different sides; their
extrinsics are calibrated to the robot base with a cube held by the gripper,
which is used only for calibration. The bench holds ordinary lab equipment:
shallow white dishes of powdered reagents, clear plastic cups, a plastic
beaker, a plastic transfer pipette standing in a lidded paper cup, and a
3D-printed scoop, stirrer, and stirrer holder. Each container is stacked on a
slip of paper with its contents written or printed on it.

\textbf{Software.} 
The system runs as two processes: the segmentation model
SAM~3~\cite{carion2026sam} runs as a separate GPU service, which the
program controlling the arm calls over the local network. The agents follow
the modular structure of PLATO~\cite{car2026plato} and consist of four
parts. An instruction parser extracts the goal, steps, reagents and
equipment from the photo of the sheet. A scene agent matches the objects
seen by the four cameras to the names in the list of bench objects. A
planner decomposes the goal into ordered sub-tasks and replans when a step
fails. Before each sub-task, a skill-call agent produces one skill call from
the measured state of the gripper; its output is constrained by a JSON
schema to a fixed set of skills. All four use GPT-4.1.
Each manipulation skill performs its own perception, grasp
analysis and motion, and checks during execution that it has reached its
target. Fig.~\ref{fig:architecture} shows which parts run once per
experiment and which run for every sub-task.

\textbf{Data flow.} A run starts from one photo of the instruction sheet.
The parser turns it into the goal and the steps, the scene agent turns two
of the camera views into the names of the objects on the bench, and the
planner turns both into an ordered list of sub-tasks. For each sub-task,
the skill-call agent is given the sub-task, the state of the gripper read
from the arm and the outcome of every earlier call, and returns one call,
for example a \texttt{pour} with its target container and a speed. The
skill fills any parameter left out with the value tested on this bench,
locates its target with the label vote of Sec.~\ref{sec:labels} and acts.
If it fails, the planner is given the failed sub-task, the sub-tasks
already done and the reason, and plans the rest again, up to three times
before the run stops.

\section{Multi-View Perception}
\label{sec:perception}

\subsection{Open-Vocabulary Segmentation and 3D Fusion}
\label{sec:seg}

The system needs only a list of the objects on the bench, not their
positions. A tool is found by a SAM~3 text prompt, such as ``white plastic
tool'' for the scoop. A labelled container is found by its category
(``white bowl'' or ``clear plastic cup'') and its label
(Sec.~\ref{sec:labels}); a transparent one also has its radius and height
listed. A locate captures a colour image and an aligned depth image from
each of the four cameras, and SAM~3 returns every instance of the prompt in
each image with its score.

Opaque objects are placed from depth. Each camera's highest-scoring mask,
if its score is at least 0.5, is back-projected into the robot base frame.
Only pixels whose depth is within the object's apparent size of the mask's
median depth are kept, which removes the table seen past the object's
edges. The largest set of cameras whose centroids all lie within 50\,mm of
each other is merged. Two cameras that disagree are both refused, and the
locate returns \emph{not found}.

Transparent containers cannot be placed from depth. Depth passes through
the wall to the table, and the depth centroids of one cup from four cameras
were up to 6\,cm apart. Each camera instead casts a ray through the
centroid of its mask. The largest set of at least two cameras whose rays
pass within 25\,mm of one point inside the container's height is kept, so
rays that only cross in mid-air are rejected. On four containers the rays
passed 6 to 11\,mm from the point. The skills receive a cylinder of the
listed size at that point.

When a category has several containers, every instance in every camera is
placed this way, and instances from different cameras at the same spot are
grouped as one container. The labels then tell these containers apart (Sec.~\ref{sec:labels}).

\begin{table}[!h]
  \caption{Label reading by label set.}
  \label{tab:label_reading}
  \centering
  \footnotesize
  \begin{tabular}{@{}lcc@{}}
    \toprule
    Label set & Right container & All 4 cameras read it \\
    \midrule
    Printed       & 40/40   & 35/40 \\
    Handwriting 1 & 40/40   & 33/40 \\
    Handwriting 2 & 40/40   & 32/40 \\
    Handwriting 3 & 40/40   & 39/40 \\
    \midrule
    Total         & 160/160 & 139/160 \\
    \bottomrule
  \end{tabular}
\end{table}

\begin{table}[!h]
  \caption{Containers found after being moved off their labels.}
  \label{tab:label_distance}
  \centering
  \footnotesize
  \begin{tabular}{@{}lcc@{}}
    \toprule
    Moved container & \makecell{Other labels\\kept} & \makecell{Other labels\\removed} \\
    \midrule
    Baking soda dish & 18/18 & 10/18 \\
    Water cup        & 15/15 & 12/15 \\
    Oil beaker       & 15/15 & 10/15 \\
    \midrule
    Total            & 48/48 & 32/48 \\
    \bottomrule
  \end{tabular}
\end{table}

\subsection{Reading Human-Written Labels (P3)}
\label{sec:labels}

Labels are placed only on containers that look identical and cannot be
told apart directly. Each such container stands on a slip of paper with its
contents written or printed on it; each tool appears once and carries no
label. Because a container's position comes from multi-view geometry
(Sec.~\ref{sec:seg}), its label only has to say what the container holds.
It needs no calibration and no fixed position or orientation, and people
can read it. It does have to be visible next to its container: a container
standing on or against its paper is read, and one with bare table between
it and the paper is not (Sec.~\ref{sec:exp_labels}).

Each instance in each camera is read on its own, from a crop of the
container and the paper it stands on, with other containers greyed out.
The VLM (GPT-4.1) chooses which of the labels on the bench is on that
paper, or none; any other reply counts as unread.
Sec.~\ref{sec:exp_labels} tests this on printed and handwritten labels
(Table~\ref{tab:label_reading}) and with containers moved off their labels
(Table~\ref{tab:label_distance}).

\begin{figure}[t]
  \centering
  \includegraphics[width=\columnwidth]{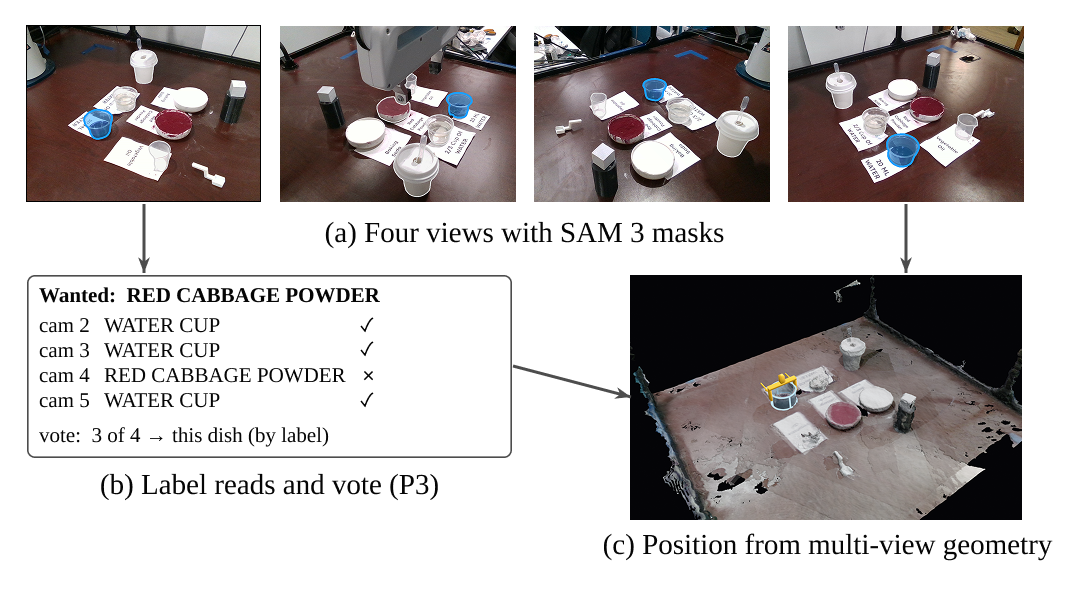}
  \caption{One locate on the real bench. (a)~SAM~3 masks for the prompt in
  each of the four camera views. (b)~The label read on the container in
  each view and the vote across cameras. (c)~The container's position from
  multi-view geometry, which the skill then uses.}
  \label{fig:perception}
\end{figure}

Containers of one category form a vote group (the two white dishes, or the
clear cups and the beaker). The labels of the group are given out, one
container each, so that the most camera reads agree. A container can then
be found by label or by elimination. It is found \emph{by label} when every
best assignment gives it the wanted label and at least one camera read that
label on it. It is found \emph{by elimination} when no camera read the
wanted label, every other label of the group is settled on a container seen
by two or more cameras, and exactly one such container is left. A tie goes
to the container nearest where it was last seen, and any other case returns
\emph{not found}. A container the arm set down itself is taken to be the
one within 50\,mm of that spot, without reading labels, because its paper
stays behind when the arm moves it. Fig.~\ref{fig:perception} shows the
camera views, the label reads and the resulting position for one locate.

\section{Manipulation Skills}
\label{sec:skills}

Every skill runs the same steps: (1)~parameters a call leaves out take
values validated on this bench; (2)~the arm moves to its home pose, out of
the cameras' view, and the skill locates its target
(Sec.~\ref{sec:perception}), reusing a kept scan when one camera shows no
change around the target; (3)~the skill plans its motion from measured
geometry, a container's rim circle and, for a held object, its shape
measured at the pick, and executes it; (4)~it checks that its moves end
near their targets, and checks the gripper width to detect a dropped
object. The put-back skill is the exception: it goes straight to the
remembered pick site without locating anything. The scans of everything a
skill handles are dropped, so the next skill locates them again (P4).
The skills and agents were first developed in a MuJoCo~\cite{6386109}
model of the bench, where they ran end to end on random container layouts;
the skills' default parameters were then set on the real bench.

\textbf{Grasping.} Every pick is a top grasp over the object's centre,
measured from its point cloud. A tool is closed on until the finger force
reaches a set limit. A thin plastic cup gives no useful force reading, so
the jaws close to its measured width less a small squeeze. A long tool
lying at an angle, such as the scoop, is turned to: the wrist rotates to
the tool's long axis, so the tool sits in the jaws the same way wherever
it lies. After closing, and again after lifting, the gripper width is
compared with the measured width, which catches a grasp that closed on
nothing, crushed the cup or lost the object in the lift. The pick also
measures what it holds, such as how far the scoop's head hangs below the
jaws and the size of a held cup, and later skills use these measurements.

\textbf{Scoop and dump.}
The scoop digs along a straight line through the
centre of the dish, as long as its head can travel without touching the
wall of the cup. The powder level is read from the depth points where the head
enters, and the head descends compliantly onto the surface and then to a
fixed depth under it, keeping a fixed gap above the floor. The scoop then
rolls from nose-down to level as it moves forward, and tilts back as it
lifts so the powder stays in. A waypoint that ends far from its target
counts as a jam, and the stroke is tried once more, higher. The dump
carries the scoop over the target cup and tips it nose-down with the
centre of its head held over the cup. The angle is capped by what the arm
can reach at that spot, and the skill fails if the cap or the measured
tip is below a minimum. Finally, it shakes the scoop and twists the wrist
to release what is left.

\textbf{Pour.} The pour takes the held cup's size from the pick and
locates the target's rim. It places the lip of the held cup just inside
the target's rim and tips toward the arm's base in steps up to the pour
angle. At each step the wrist position is recomputed so the lip stays
over the target and the cup's body clears its rim. After each step the
measured tilt is compared with the commanded one; a step that stops short
is tried again more slowly, and if it still stops short the pour fails.
The cup is held at the final angle, set upright in place and lifted
clear.

\textbf{Stir and tool return.} The stir holds the stirrer straight down,
locates the target cup and measures its rim, then lowers the rod in small
steps to its depth, stopping early if it touches anything. Each
revolution is traced as one continuous motion on a circle sized to the
measured opening, so the rod cannot scrape the wall, and the skill checks
before, during and after that the stirrer is still held. A tool is put
back where it was picked up, and the stirrer is lowered back into its
holder.

\textbf{Pipette.} \texttt{pick\_pipette} takes the pipette out of its cup
by the bulb, squeezed flat, dips the tip into the source and opens the
jaws a little to draw liquid up; \texttt{dispense} releases it as drops.

\section{From Instruction Sheet to Skill Calls}
\label{sec:planning}

\textbf{Instruction parsing.} The parser sends the photo of the sheet to
the VLM, which returns the goal, the ordered steps, the reagents with
their quantities and the equipment as JSON. It is told to keep each
step's own wording for which container and how much. Without that line it
read ``the empty clear cup'' as ``the beaker'' in 2 of 3 reads; with it,
it read the step correctly in 3 of 3.

\textbf{Scene grounding.} The list of bench objects is part of the bench
setup: each object has a name, its label if it has one, and a short note,
for example that a cup already holds its water. The scene agent looks at
two of the four camera views and names the objects on the bench, choosing
each name from this list, so every name it gives is one a skill can
locate. It is told what the objects are called, not where they are.

\textbf{Planning and skill calls.} The planner turns the goal, the steps
and the named objects into an ordered list of sub-tasks. It is shown the
fixed set of skills and the rules for combining them, for example that a
scoop only fills the bowl and needs a dump to deliver anything, and that
numbered steps on the sheet keep their order. For each sub-task, the
skill-call agent returns one call under a strict JSON schema: a skill
from the fixed set, its parameters, a speed and a short reason. A
sub-task it cannot express as a skill is reported as infeasible, rather
than mapped to a nearby skill, and goes back to the planner. The model is
not asked for what it cannot know: the shape of the scoop's head comes
from its CAD model, and measurements taken at a pick, such as where the
scoop sits in the jaws, are passed on to the calls that use the tool.

\textbf{Execution and recovery.} Each call's parameters, outcome, timing
and the values the skill measured are written to a run record. A failed
skill goes back to the planner, as described in
Sec.~\ref{sec:overview}. An operator can step through a run, confirming
each call before it runs, and a stopped run can be resumed from its
record at a chosen sub-task without repeating the steps already done.

%
\section{Experiments}
\label{sec:experiments}


\subsection{Setup and Protocols}
\label{sec:exp_setup}
We evaluate \sysname{} on two protocols, Magic Beaker and jellyfish,
each given to the system only as a photo of its printed sheet. Both
protocols scoop powders, pour liquids and stir; jellyfish also adds drops
with the pipette. Both come from a National Geographic Kids chemistry kit:
the sheets are pages of its booklet, and the reagents, cups, beaker,
pipette and scoops are those of the kit. In Magic Beaker, citric acid and
baking soda are each dissolved in a cup of water, giving an acid and a
base, and red cabbage powder is dissolved in the beaker as a pH indicator.
The indicator is poured into both cups, turning the acid red and the base
blue, and the two are then poured into an empty cup, where the liquid
fizzes and turns purple. The sheet pours the two at the same time; the
robot has one arm and pours them one after the other. In jellyfish, baking
soda is dissolved in a cup of water, vegetable oil is poured on top, and
red cabbage indicator mixed in a second cup is dropped onto the oil with
the pipette; the purple drops sink through the oil and turn blue in the
base below. The planner decides the number of sub-tasks itself, so it
varies from run to run: about 30 for Magic Beaker and about 20 for
jellyfish.

The bench is set up the way a chemistry lab would be. Every container of
a liquid or a powder stands on a paper label naming its contents, as a lab
labels its reagents. The tools (scoop, stirrer and pipette) carry no
label, since a chemist knows them by sight, and the system has to as well.
The reagents come already measured into their own containers, much like
mise en place in a kitchen, so the robot transfers and mixes them but does
not measure out amounts. Neither protocol depends on precise amounts; we
return to this in Sec.~\ref{sec:limitations}.

Each protocol was run 5 times. A run counts as a success when every
sub-task is carried out with no human intervention and the expected
visible result appears. Between runs the containers and tools were
shifted slightly by hand, but the arrangement stayed the same and no two
objects swapped places.

\subsection{End-to-End from the Printed Protocol (P1, P2)}
\label{sec:exp_e2e}
Magic Beaker succeeded in 5 of 5 runs and jellyfish in 4 of 5
(Table~\ref{tab:e2e}). A Magic Beaker run took about 30 minutes and a
jellyfish run about 17. Every successful run ended with the result its
sheet describes: in Magic Beaker the mixed liquid fizzed and turned
purple, and in jellyfish the indicator drops sank through the oil and
turned blue.

The one failed run was a jellyfish run. Every sub-task went as planned
until the final steps, where the pipette is picked up to drop the
indicator onto the oil. There the system took the scoop for the pipette
and grasped the scoop instead. Tools carry no label, so the label vote
(Sec.~\ref{sec:labels}) cannot catch this: the pipette is found only by
the text prompt the system gives SAM~3 for it (Sec.~\ref{sec:seg}). A
pipette is something a chemist knows by sight, so it is left unlabelled
on purpose, and the system has to recognise it the same way. We expect a
stronger VLM to do better at identifying the pipette and choosing the
prompt for it.

\begin{table}[!h]
  \caption{End-to-end runs from the photographed protocol, with no human
  intervention.}
  \label{tab:e2e}
  \centering
  \footnotesize
  \begin{tabular}{@{}lccc@{}}
    \toprule
    Protocol & Runs & Success & Time (min) \\
    \midrule
    Magic Beaker & 5 & 5/5 & 30 \\
    Jellyfish    & 5 & 4/5 & 19 \\
    \bottomrule
  \end{tabular}
\end{table}

\subsection{Identifying Containers by Their Labels (P3)}
\label{sec:exp_labels}

Both label experiments use the same bench. Four containers stand in a
square in front of the arm, each on its paper label: two identical white
dishes (red cabbage powder and baking soda), a clear plastic cup (10\,ml
water) and a plastic beaker (vegetable oil). The VLM chooses among
exactly these four label strings. Only the cameras are used; the arm does
not move. Each locate is the same call a skill makes, and every memory is
cleared before it (kept scans, last known position and tracked objects),
so no locate is helped by the one before. Each locate is settled by label
or by elimination, or returns \emph{not found} (Sec.~\ref{sec:labels}).
The ground truth
is each container's position located with label reading switched off and
named by its corner of the square; three repeats agreed within 0.4\,mm. A
locate with all four cameras took 14.1\,s on average (13.2 to 18.1\,s).

\textbf{Label legibility.} The labels were swapped between four sets:
printed, and handwritten by three different people. In each set every
container was located 10 times (Table~\ref{tab:label_reading}). All 160
locates found the right container, with a mean offset of 0.2\,mm from the
ground truth and at most 1.1\,mm. The vote across cameras is needed for
this. In 21 of the 160 locates one camera read another label on the right
container: camera 3 in 11, always on the red cabbage dish and on both
printed and handwritten paper, probably because it sees that paper
sideways at a low angle; camera 4 in 9, all on the beaker and 8 of them in
one handwriting set; and camera 2 once. No locate had more than one camera
misreading, so the other three outvoted it every time. Used alone, camera
3 would have found at most 149 of the 160 containers.

\textbf{Distance from the label.} One container at a time (the baking
soda dish, the water cup or the beaker) was moved away from its label to
5 or 6 spots between 2 and 44\,cm, and located 3 times at each spot, 96
locates in all. The distance is how far its centre moved, measured by
hand. Each container was tested twice: with the other three labels left on
the bench, and with those labels removed while their containers stayed
(Table~\ref{tab:label_distance}). None of the 96 locates returned the
wrong container; every miss was \emph{not found}. With the other labels
present no locate failed, because whenever the moved container's own
label was not read, elimination found it (10 locates). With the other
labels removed, 16 of 48 locates were not found. The distance of the
centre does not predict whether the label is read. The water cup was read
by all four cameras 18\,cm from its spot, standing against the edge of its
paper, and was not found at 16\,cm with bare table between it and the
paper. The baking soda dish was not read at 6\,cm, where it stood on top
of its own text. What decides the read is whether the label is still in
view beside the container, that is, whether the container stands on or
against its paper. We saw this in a few frames and did not measure the gap
from the container to its paper or the direction of the move.

Two results need care. First, with the other labels removed, elimination
still settled 2 of the 48 locates, because the VLM named a container
from what it holds: one camera read the red cabbage dish as ``red cabbage
powder'' from its red powder, and another read the water cup as ``10\,ml
water''. The rows without the other labels are therefore not a clean
condition with no other labels. Second, two far reads counted as by label
rest on a single camera and were not checked against the frames: the
baking soda dish at 44\,cm (2 of 3 locates) and at 18\,cm without the
other labels. They may come from the contents of the dish and not from its
label. Elimination was not switched off in the code for the phase without
the other labels. Removing those labels takes away the moved container's
only elimination partner without changing the system, so the test runs the
same code as a real run.
\subsection{Robustness to People Moving Things (P4)}
\label{sec:exp_rescan}
We have not yet evaluated this property with a person moving objects
during a run. This section describes how the system is built to handle
it. Every skill locates its target again from the home pose right before
it acts (Sec.~\ref{sec:skills}), so a container a person has moved
between two steps is found at its new spot. A kept scan is reused only if
no skill has touched the object since and one camera's current view of
the object matches the view the scan came from; any visible change there,
such as a hand moving the object, forces a full locate. Identity comes
from the label vote and not from where a container stood before, so two
identical cups that a person swaps, each with its label, are still told
apart. A container the arm set down itself has no label under it; it is
taken to be the one within 50\,mm of where the arm left it, and if none
is there, the labels are read instead. Tools carry no label and are found
by their text prompt at every pick, so a moved tool is found the same
way. If a person takes an object out of the gripper, the gripper width
shows it, the skill fails, and the planner replans from the measured
gripper state.

Some cases are not covered. Each skill locates once before it moves, so
an object moved while a skill is acting is not seen until the next
skill. The put-back skill returns an object to where it was picked up
without looking, so anything placed on that spot in the meantime is in
the way. And because the label is the identity, a label moved to another
container takes the identity with it, as it would for a person reading
the bench.


\subsection{Execution Adapted to What Is Handled (P5)}
\label{sec:exp_speed}
This property has not been evaluated on its own yet either; this section
describes how it works. A person carries a full cup slowly and an empty
one quickly, and so does \sysname{}. Every skill call carries a speed from
0 to 100 that scales the duration of every motion in the skill, from the
pace each skill was tuned at (0, the slowest) down to 0.4 of that time at
100. Holds and waits keep their length. The whole call runs at one speed,
so the skill-call agent chooses it for the call's most delicate moment,
following a short guide in its prompt: 0 for liquid in an open container,
20 for a scoop carrying powder, 30 for stirring, 60 for an empty tool and
80 with nothing held. One fixed pace would either spill liquid or waste
time on every empty move. The other parameters are set the same way: the
agent chooses the pour angle from the wording of the step, a fixed rule
sets it when the same container is poured from or into more than once,
and the scoop depth comes from the measured powder surface. All ten
end-to-end runs of Sec.~\ref{sec:exp_e2e} used the speeds the agent chose.

Speed only shortens the motions, and much of a run is spent looking. Every
skill locates its target again before it acts, which is what lets it find
an object a person has moved (Sec.~\ref{sec:exp_rescan}). A locate reads
the four cameras one at a time. Opening all four RealSense streams at
once overloaded the USB controller of the robot's computer, and the GPU
does not have the memory to run SAM~3 on all four views together, so
each camera is started, read and stopped before the next, and SAM~3 and
the label reads then run on each view in turn. A locate with all four
cameras took 14.1\,s on average in the label experiments
(Sec.~\ref{sec:exp_labels}). Reusing a scan when nothing around the
target has changed saves some of these, but with about 30 sub-tasks in a
Magic Beaker plan the locates are a large part of why a run takes about
20 minutes (Table~\ref{tab:e2e}). A computer that can stream all four cameras and segment
them at once would remove most of this wait.


\section{Limitations}
\label{sec:limitations}

The clearest limitation is amounts. The reagents come already measured
(Sec.~\ref{sec:exp_setup}), both protocols work whatever the exact
amounts, and the robot does not weigh or otherwise measure what it
transfers. Our results therefore say nothing about how accurately
\sysname{} could deliver a set amount, which most real protocols need.

The evaluation is also narrow. It covers one bench, one arm and two
protocols from a children's kit, and the end-to-end runs kept one
arrangement of the bench, shifting objects only slightly between runs.
The arm's joint limits restrict where on the bench some skills can be
carried out; the dump, for instance, cannot tip as far over a cup near
the edge of its reach. Properties P4 and P5 are built into the system,
but neither has yet been tested on its own (Secs.~\ref{sec:exp_rescan}
and~\ref{sec:exp_speed}).

Perception has its own gaps. Unlabelled tools are found by a text prompt
alone, which is how the one failed run went wrong
(Sec.~\ref{sec:exp_e2e}). The cameras look down on the bench, so the
underside of the scoop's head is hidden and its measured depth below the
grasp is only a lower bound. Every locate reads the four cameras one
after another, which makes a run slow (Sec.~\ref{sec:exp_speed}). And the
system depends on a hosted VLM, so it needs a network connection and is
tied to the model's behaviour.

\section{Conclusion}
\label{sec:conclusion}

We presented \sysname{}, a system that carries out a chemistry experiment
on a robot arm from a photo of its printed instruction sheet, on a bench
set up for people. Containers stand loosely on handwritten or printed
labels, which the system reads in four camera views and combines by a
vote, and it locates its target again before every step. It completed 9
of 10 runs of two protocols with no human intervention and identified
every container in 160 label-reading locates. Next, we will test it with
people moving objects during a run, compare the speeds it chooses
against fixed ones, and move to protocols that need measured amounts.


\balance
\bibliographystyle{IEEEtran}
\bibliography{references}

\end{document}